\documentclass[ASNA,twocolumn]{USG}
\usepackage{placeins}
\graphicspath{{./images/}}
\definecolor{revieworange}{HTML}{C65D00}

\hypersetup{hidelinks,
 pdftitle={ForensicJev: Adapting Jev-style decision models for image forgery detection and source attribution},
 pdfauthor={},pdfsubject={Forensic task adaptation; Electronics Letters manuscript}}

\makeatletter
\patchcmd{\@maketitle}{\includegraphics[width=26.5mm]{Wiley_logo.eps}}{}{}%
 {\PackageError{manuscript}{Publisher logo removal failed}{Check USG.cls version.}}
\patchcmd{\@maketitle}{\smash{\@journalfont\@journal}}{}{}%
 {\PackageError{manuscript}{Journal heading removal failed}{Check USG.cls version.}}
\patchcmd{\@maketitle}{\vskip24.36mm}{}{}%
 {\PackageError{manuscript}{Logo-space removal failed}{Check USG.cls version.}}
\patchcmd{\@maketitle}{\@@articletype}{}{}%
 {\PackageError{manuscript}{Article badge removal failed}{Check USG.cls version.}}
\patchcmd{\@maketitle}{\vskip8.6mm}{}{}%
 {\PackageError{manuscript}{Badge-space removal failed}{Check USG.cls version.}}
\makeatother

\makeatletter
\patchcmd{\head@headings@def}{\vskip3.5mm}%
 {\hsize\textwidth\vskip3.5mm}{}%
 {\PackageError{manuscript}{Running footer compatibility patch failed}{Check USG.cls version.}}
\makeatother

\makeatletter
\renewcommand{\oddhead@titlepage@info}{}
\renewcommand{\evenhead@titlepage@info}{}
\renewcommand{\oddfoot@titlepage@info}{%
 \head@headings@def{\hfill\pagenumfont\thepage\ of \@LastPg}}
\makeatother

\makeatletter
\renewcommand{\evenhead@headings@info}{\head@headings@def{\evenpage@headings@def\hfill}}
\makeatother

\RemoveFromHook{shipout/after}[USG]
\newcommand\blfootnote[1]{\begingroup\renewcommand\thefootnote{}\footnote{#1}\addtocounter{footnote}{-1}\endgroup}

\begin{document}
\journal{Electronics Letters}
\articletype{LETTER}
\title{ForensicJev: Adapting Jev-style decision models for image forgery detection and source attribution}
\author[1]{Yaqi Li$^{1}$}
\author[1]{Yabin Wang$^{1}$}
\author[1]{Xiaopeng Hong$^{1}$}
\address[1]{\orgname{Harbin Institute of Technology}}
\corres{Xiaopeng Hong (\email{hongxiaopeng@ieee.org})}
\authormark{Y. Li et al.}
\titlemark{ForensicJev: Forgery detection and source attribution}
\copyyear{}
\volume{}
\articledoi{}
\startpage{1}
\copyright{}

\abstract[ABSTRACT]{
This paper introduces ForensicJev, the first Jev-style framework to unify image forgery detection and source attribution. We formulate both tasks as choice tasks, casting forgery detection as binary choice and source attribution as variable-length multiple choice. Under this unified formulation, the backbone parameters and classification head are shared across both tasks. To prevent candidate-position bias, we design a candidate shuffling mechanism during training. We further construct a benchmark for forgery detection and source attribution covering 85 forgery sources. On this benchmark, ForensicJev outperforms specialist detectors and performs on par with or surpasses visual foundation models across both tasks, achieving the highest detection average precision of 97.39\% and the highest 85-way source attribution accuracy of 62.32\%.}
\maketitle
\section{Introduction}
\label{sec:introduction}
\blfootnote{Emails: Yaqi Li (\email{25S103223@stu.hit.edu.cn}), Yabin Wang (\email{wang-yabin@outlook.com}).}Image forgery detection, which determines whether an image is authentic, and source attribution, which identifies the generative model or editing pipeline behind synthetic or manipulated content, are two central tasks in image forensics. A unified framework that addresses both tasks can provide a more comprehensive forensic analysis of an image. Existing methods primarily rely on visual-only frameworks, where specialist detectors~\citep{wang2020cnn,tan2024npr,ojha2023universal,yan2025effort} and visual foundation models~\citep{simeoni2025dinov3,tschannen2025siglip2} have established strong visual baselines for image forensics. Beyond these visual backbones, Jev-style multimodal decision models have recently emerged as a promising approach to unifying both tasks~\citep{typesafe2026jev,startlux2026}. However, their application to image forensics remains unexplored. To fill this gap, we adapt Jev-style decision models to both tasks.

In this paper, we propose ForensicJev, the first Jev-style framework to unify image forgery detection and source attribution, which adapts StartLux-Decision-v1-4B~\citep{startlux2026} by jointly fine-tuning its visual encoder and language model with LoRA~\citep{hu2021lora}, while incorporating an auxiliary visual branch over multi-layer visual features for source attribution. Within this framework, we formulate both tasks as choice tasks through task prompts, casting forgery detection as binary choice and source attribution as variable-length multiple choice. Under this unified formulation, the backbone parameters and classification head are shared across both tasks. During training, we shuffle candidate order to prevent candidate-position bias and vary list length to support variable-length multiple choice, encouraging the model to rely on visual evidence and task prompts rather than fixed candidate configurations. We further construct a benchmark for forgery detection and source attribution covering 85 forgery sources. ForensicJev outperforms specialist detectors and performs on par with or surpasses visual foundation models across both tasks. For forgery detection, it achieves 90.31\% accuracy and the highest average precision of 97.39\%. For source attribution, it achieves the highest accuracy of 62.32\% and 59.58\% macro-F1.
\section{Method}
\label{sec:methodology}

\begin{figure}[!t]
\centering
\includegraphics[width=\linewidth]{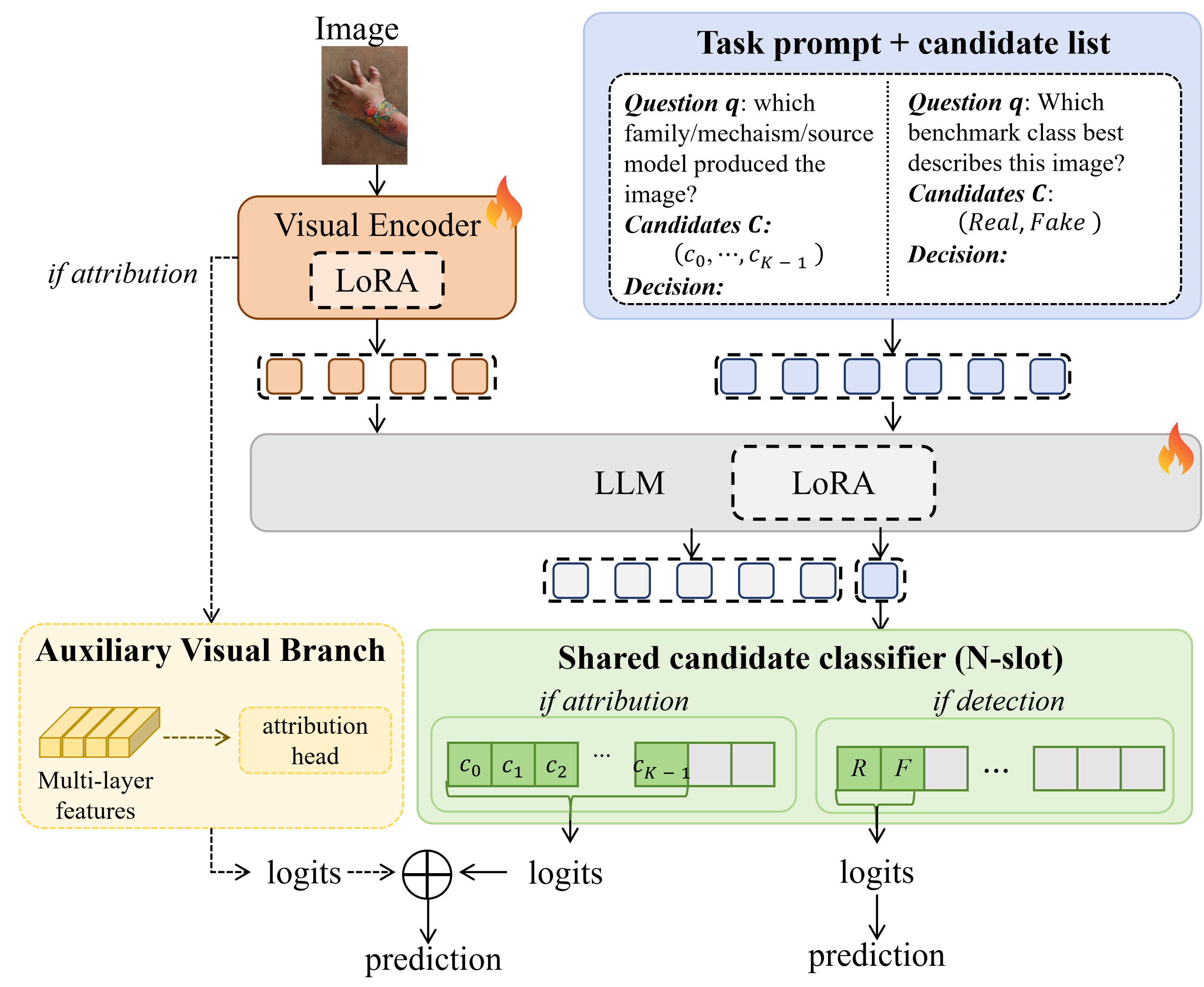}
\caption{Overview of the ForensicJev framework. Given an input image, a task prompt, and a candidate list of length $K$, the LoRA-adapted visual encoder and language model jointly encode the inputs, and the shared $N$-slot candidate classifier scores the first $K$ active slots from the final hidden state. For forgery detection, predictions rely solely on the shared classifier. For source attribution, the auxiliary branch extracts multi-layer visual features and adds candidate-aligned logits to the active slots before softmax.}
\label{fig:method-overview}
\end{figure}

ForensicJev adapts StartLux-Decision-v1-4B to image forgery detection and source attribution (Figure~\ref{fig:method-overview}). Given an input image, a task prompt, and a candidate list, ForensicJev selects an answer by estimating the probability of each candidate. The visual encoder encodes the input image into visual tokens, and the language model jointly processes these tokens with the task prompt and candidate text. The hidden state of the last input token is mapped to candidate scores by a classifier shared across both tasks, with an auxiliary visual branch incorporating multi-layer visual features to provide additional scores for source attribution. To adapt the model to image forensics, we jointly fine-tune the visual encoder and the language model with LoRA~\citep{hu2021lora}. In the remainder of this section, Section~\ref{sec:prompts} details task prompt design and candidate list construction, Section~\ref{sec:classifier} presents the shared candidate classifier, and Section~\ref{sec:aux-branch} describes the auxiliary visual branch and joint training.
\subsection{Task prompts and candidate construction}
\label{sec:prompts}
\noindent\textbf{Task prompts.}
ForensicJev unifies binary forgery detection and source attribution by formulating both tasks as choice tasks over a numbered candidate list, rather than free-form text generation: detection is cast as binary choice, while source attribution is cast as variable-length multiple choice over 85 forgery sources. Because the shared candidate classifier scores candidates solely by their list positions, each task prompt must specify both the target task and the semantics of each candidate. Each prompt instructs the model to evaluate visual evidence and select one numbered candidate. For detection, the task prompt asks ``Which benchmark class best describes this image?'', where Real denotes authentic content without generation or content manipulation, whereas Fake covers full generation and content manipulation, including local editing, face synthesis or swapping, and compositing. The input image is paired with the following prompt structure, with candidate descriptions abbreviated for brevity:
\begin{flushleft}\ttfamily
Question: Which benchmark class best describes this image?\\
Candidates:\\
\mbox{[0]} Real: An authentic image \ldots\\
\mbox{[1]} Fake: A synthetic or\\content-manipulated image \ldots\\
Decision:
\end{flushleft}
For source attribution, the task prompt assumes synthetic or manipulated content and asks which specific generative model or editing pipeline produced the image. This prompt focuses on the creation process rather than the depicted subject, using the predefined category names of the forgery sources as candidates.

\noindent\textbf{Candidate construction.}
To prevent candidate-position bias and enforce prompt-conditioned matching, we randomly shuffle every candidate list during training and update its target index $y$ accordingly. Furthermore, to support variable-length multiple choice rather than restricting the model to a fixed candidate list length, we apply variable-length candidate sampling to source attribution, retaining the complete 85-candidate list for most updates while otherwise sampling shorter subsets that pair the ground-truth source with randomly drawn distinct negatives. During validation and testing, both tasks are evaluated using canonical ordering over their complete candidate lists.

\subsection{Shared candidate classifier}
\label{sec:classifier}
In ForensicJev, a single classification head is shared across binary detection and source attribution, accommodating candidate lists of varying lengths. Given an input image $x$, a task prompt $q$, and a candidate list $C=(c_0,\ldots,c_{K-1})$ of length $K$, let $h(x,q,C)$ denote the final hidden state at the last valid input position in the adapted language model. A fixed $N$-slot linear classifier, where $N$ is chosen to be at least the maximum number of candidates across tasks, maps this state to logits
\begin{equation}
u=Wh(x,q,C)+b\in\mathbb{R}^{N}.
\end{equation}
For a given task and its candidate list $C$ of length $K$, only the first $K$ slots are activated, while the remaining $N-K$ slots are masked. Each active slot aligns with the corresponding candidate position in $C$ rather than a fixed semantic class.

\subsection{Auxiliary visual branch and joint training}
\label{sec:aux-branch}
\noindent\textbf{Auxiliary visual branch.}
To preserve fine-grained artifact cues that can be attenuated across the language backbone, ForensicJev incorporates an auxiliary multi-layer visual branch for source attribution. This branch extracts patch features from four intermediate blocks of the adapted visual encoder, applies mean and max pooling to each, concatenates the pooled representations, and passes them through layer normalization and a linear attribution head. To match the shuffled and variable-length candidate lists described in Section~\ref{sec:prompts}, the resulting category scores are indexed by candidate identity and reordered into $v(C)$ before being added to the active slots of $u$. For each active slot, the final logit is
\begin{equation}
z_j=\begin{cases}
u_j, & \text{detection},\\
u_j+v_j(C), & \text{attribution}.
\end{cases}
\end{equation}

\noindent\textbf{Joint training.}
All tasks are optimized jointly through the same choice objective over the final logits $z$:
\begin{equation}
p(c_j\mid x,q,C)=\frac{\exp z_j}{\sum_{k=0}^{K-1}\exp z_k},\quad
\mathcal{L}=-\log p(c_y\mid x,q,C),
\end{equation}
where $p(c_j\mid x,q,C)$ denotes the predicted probability of selecting the $j$-th candidate over the $K$ active slots, and $\mathcal{L}$ is the cross-entropy loss with respect to the ground-truth target position $y$. Crucially, $y$ indexes a position within the presented candidate list rather than a fixed class identifier, allowing the same objective and shared classifier to optimize both binary detection and source attribution without architectural modification.
\section{Results and discussion}
\label{sec:experiments}
\textbf{Dataset.} To evaluate forgery detection and source attribution within a unified benchmark, we construct a dataset combining authentic images with globally generated and locally edited forgeries. The benchmark covers 85 forgery sources. The training, validation, and test splits contain 41,650, 5,376, and 10,702 images, respectively, with balanced real and fake counts per split.

\textbf{Experimental setup.} All evaluated systems use the same data splits. As indicated in Table~\ref{tab:detection-context}, four specialist detectors~\citep{wang2020cnn,tan2024npr,ojha2023universal,yan2025effort} train with detection-only supervision (D), whereas four conventional pretrained backbones~\citep{he2016resnet,tan2019efficientnet,liu2021swin,liu2022convnext}, two visual foundation models (DINOv3 and SigLIP2~\citep{simeoni2025dinov3,tschannen2025siglip2} with visual LoRA and final-layer classifiers), and ForensicJev (along with its language-only LoRA ablation, ForensicJev (Lang-LoRA)) use joint detection--attribution supervision (D+A). Built on StartLux-Decision-v1-4B, ForensicJev uses rank 16, scaling 32, and a dropout rate of 0.05 for the language LoRA adapter, and rank 8, scaling 16, and no dropout for the visual LoRA adapter. During source attribution training, the full 85-candidate list is presented with probability 0.8 and a subset of size $K \in [2, 84]$ is sampled otherwise, while the auxiliary visual branch pools patch features from blocks 6, 12, 18, and 24 of the 24-block visual encoder. ForensicJev is trained using AdamW~\citep{loshchilov2019decoupled} with zero weight decay, learning rates of $10^{-4}$ for LoRA adapters and $5\times10^{-4}$ for linear heads, and 5\% linear warmup followed by cosine decay. Detection uses accuracy (ACC), average precision (AP), and fake-F1, while attribution uses ACC and macro-F1 on fake images.
\begin{table}[!t]
\centering
\caption{Image forgery detection results on the test set (\%). D and A denote detection and attribution supervision, respectively. Pretrained StartLux uses its native answer-scoring interface. Best and second-best scores are bolded and underlined.}
\label{tab:detection-context}
\begin{tabularx}{\linewidth}{@{}>{\raggedright\arraybackslash}Xcrrr@{}}
\toprule
System & Supervision & ACC & AP & Fake-F1\\
\midrule
\multicolumn{5}{l}{\textit{Pretrained StartLux}}\\
StartLux, pretrained & -- & 58.04 & 62.33 & 50.42\\
\midrule
\multicolumn{5}{l}{\textit{Specialist detectors}}\\
CNNSpot & D & 71.35 & 85.93 & 75.49\\
NPR & D & 62.73 & 81.69 & 70.39\\
UniFD & D & 64.46 & 77.92 & 70.01\\
Effort & D & 86.91 & 95.15 & 87.46\\
\midrule
\multicolumn{5}{l}{\textit{Conventional backbones}}\\
ResNet-50 & D+A & 84.67 & 93.31 & 83.93\\
EfficientNet-B0 & D+A & 85.64 & 94.12 & 85.50\\
Swin-T & D+A & 85.59 & 93.06 & 84.84\\
ConvNeXt-T & D+A & 84.78 & 92.86 & 84.83\\
\midrule
\multicolumn{5}{l}{\textit{Visual foundation models}}\\
DINOv3 & D+A & \textbf{90.55} & \underline{97.33} & \textbf{90.46}\\
SigLIP2-SO400M & D+A & 89.34 & 96.35 & 88.92\\
\midrule
\multicolumn{5}{l}{\textit{ForensicJev (StartLux adaptation)}}\\
ForensicJev (Lang-LoRA) & D+A & 88.03 & 96.44 & 87.36\\
ForensicJev & D+A & \underline{90.31} & \textbf{97.39} & \underline{89.93}\\
\bottomrule
\end{tabularx}
\end{table}

\textbf{Detection effectiveness.} Table~\ref{tab:detection-context} demonstrates that task-specific adaptation enables effective forensic detection from the Jev-style interface. Relative to pretrained StartLux under its native scoring interface, ForensicJev raises detection accuracy by over 32 percentage points to 90.31\% and achieves the highest AP overall at 97.39\%. Overall, ForensicJev outperforms all specialist detectors, conventional backbones, and SigLIP2 on every metric, while matching DINOv3 in ACC and fake-F1. 
\begin{table}[!t]
\centering
\caption{Source attribution results on fake images (\%). All systems use joint detection and attribution supervision. Best and second-best scores are bolded and underlined.}
\label{tab:attribution-context}
\begin{tabularx}{\linewidth}{@{}>{\raggedright\arraybackslash}Xrr@{}}
\toprule
System & ACC & Macro-F1\\
\midrule
\multicolumn{3}{l}{\textit{Conventional backbones}}\\
ResNet-50 & 57.80 & 54.91\\
EfficientNet-B0 & 59.19 & 57.24\\
Swin-T & 58.61 & 55.04\\
ConvNeXt-T & 59.11 & 55.43\\
\midrule
\multicolumn{3}{l}{\textit{Visual foundation models}}\\
DINOv3 & \underline{61.00} & \textbf{59.89}\\
SigLIP2-SO400M & 55.73 & 53.36\\
\midrule
\multicolumn{3}{l}{\textit{ForensicJev (StartLux adaptation)}}\\
ForensicJev (Lang-LoRA) & 51.00 & 47.19\\
ForensicJev & \textbf{62.32} & \underline{59.58}\\
\bottomrule
\end{tabularx}
\end{table}

\textbf{Source attribution effectiveness.} Beyond binary detection, the framework transfers effectively to source attribution (Table~\ref{tab:attribution-context}), achieving the highest accuracy at 62.32\% and 59.58\% macro-F1, closely matching DINOv3 at 59.89\%, while surpassing SigLIP2 and conventional backbones on both metrics.

\textbf{Role of visual adaptation.} Comparing ForensicJev with ForensicJev (Lang-LoRA) across Tables~\ref{tab:detection-context} and~\ref{tab:attribution-context} highlights the importance of adapting visual representations for forensics. For forgery detection, adding visual LoRA to language-only adaptation improves every metric, raising ACC by 2.28 percentage points, AP by 0.95 percentage points, and fake-F1 by 2.57 percentage points. For source attribution, jointly incorporating visual LoRA and the auxiliary multi-layer visual branch yields substantial gains, lifting accuracy by 11.32 percentage points and macro-F1 by 12.39 percentage points. These gains confirm that visual LoRA and multi-layer feature fusion are key to forensic discrimination.
\section{Conclusion}
\label{sec:conclusion}
In this paper, we presented ForensicJev, the first Jev-style framework to unify image forgery detection and source attribution, casting detection as binary choice and source attribution as variable-length multiple choice. In ForensicJev, the backbone parameters and classification head are shared across both tasks, with the visual encoder and language model jointly fine-tuned using LoRA, while candidate shuffling prevents candidate-position bias and variable-length sampling supports varying candidate lists. We further constructed a benchmark for forgery detection and source attribution covering 85 forgery sources. Experiments show that ForensicJev outperforms specialist detectors and performs on par with or surpasses visual foundation models across both tasks, achieving the highest detection AP and source attribution accuracy. These results demonstrate the potential of Jev-style decision models as a unified framework for image forensics.
\FloatBarrier
\label{page:main-end}
\bibliography{references}

\begin{thebibliography}{14}
\providecommand{\natexlab}[1]{#1}
\providecommand{\url}[1]{\texttt{#1}}
\expandafter\ifx\csname urlstyle\endcsname\relax
  \providecommand{\doi}[1]{doi: #1}\else
  \providecommand{\doi}{doi: \begingroup \urlstyle{rm}\Url}\fi

\bibitem[\protect\citeauthoryear{Wang, Wang, Zhang, Owens, and Efros}{Wang et~al.}{2020}]{wang2020cnn}
Sheng-Yu Wang, Oliver Wang, Richard Zhang, Andrew Owens, and Alexei~A. Efros. 2020.
\newblock ``{CNN}-Generated Images Are Surprisingly Easy to Spot... for Now.''   In {\it Proceedings of the IEEE/CVF Conference on Computer Vision and Pattern Recognition}.
\newblock \url{https://arxiv.org/abs/1912.11035}.

\bibitem[\protect\citeauthoryear{Tan, Liu, Zhao, Wei, Gu, Liu, and Wei}{Tan et~al.}{2024}]{tan2024npr}
Chuangchuang Tan, Huan Liu, Yao Zhao, Shikui Wei, Guanghua Gu, Ping Liu, and Yunchao Wei. 2024.
\newblock ``Rethinking the Up-Sampling Operations in {CNN}-based Generative Network for Generalizable Deepfake Detection.''   In {\it Proceedings of the IEEE/CVF Conference on Computer Vision and Pattern Recognition},   28130--28139.
\newblock \url{https://arxiv.org/abs/2312.10461}.

\bibitem[\protect\citeauthoryear{Ojha, Li, and Lee}{Ojha et~al.}{2023}]{ojha2023universal}
Utkarsh Ojha, Yuheng Li, and Yong~Jae Lee. 2023.
\newblock ``Towards Universal Fake Image Detectors that Generalize Across Generative Models.''   In {\it Proceedings of the IEEE/CVF Conference on Computer Vision and Pattern Recognition},   24480--24489.
\newblock \url{https://arxiv.org/abs/2302.10174}.

\bibitem[\protect\citeauthoryear{Yan, Wang, Jin, Zhang, Liu, Chen, Yao, Ding, Wu, and Yuan}{Yan et~al.}{2025}]{yan2025effort}
Zhiyuan Yan, Jiangming Wang, Peng Jin, Ke-Yue Zhang, Chengchun Liu, Shen Chen, Taiping Yao, Shouhong Ding, Baoyuan Wu, and Li~Yuan. 2025.
\newblock ``Orthogonal Subspace Decomposition for Generalizable {AI}-Generated Image Detection.''   In {\it Proceedings of the 42nd International Conference on Machine Learning}, Vol 267 of {\it Proceedings of Machine Learning Research},   70268--70288.
\newblock \url{https://proceedings.mlr.press/v267/yan25b.html}.

\bibitem[\protect\citeauthoryear{Sim{\'e}oni, Vo, Seitzer, Baldassarre, Oquab, Jose, Khalidov, Szafraniec, Yi, Ramamonjisoa, Massa, Haziza, Wehrstedt, Wang, Darcet, Moutakanni, Sentana, Roberts, Vedaldi, Tolan, Brandt, Couprie, Mairal, J{\'e}gou, Labatut, and Bojanowski}{Sim{\'e}oni et~al.}{2025}]{simeoni2025dinov3}
Oriane Sim{\'e}oni, Huy~V. Vo, Maximilian Seitzer, Federico Baldassarre, Maxime Oquab, Cijo Jose, Vasil Khalidov, Marc Szafraniec, Seungeun Yi, Micha{\"e}l Ramamonjisoa, Francisco Massa, Daniel Haziza, Luca Wehrstedt, Jianyuan Wang, Timoth{\'e}e Darcet, Th{\'e}o Moutakanni, Leonel Sentana, Claire Roberts, Andrea Vedaldi, Jamie Tolan, John Brandt, Camille Couprie, Julien Mairal, Herv{\'e} J{\'e}gou, Patrick Labatut, and Piotr Bojanowski. 2025.
\newblock ``{DINOv3}.''   {\it arXiv preprint arXiv:2508.10104\/}.
\newblock \url{https://arxiv.org/abs/2508.10104}.

\bibitem[\protect\citeauthoryear{Tschannen, Gritsenko, Wang, Naeem, Alabdulmohsin, Parthasarathy, Evans, Beyer, Xia, Mustafa, H{\'e}naff, Harmsen, Steiner, and Zhai}{Tschannen et~al.}{2025}]{tschannen2025siglip2}
Michael Tschannen, Alexey Gritsenko, Xiao Wang, Muhammad~Ferjad Naeem, Ibrahim Alabdulmohsin, Nikhil Parthasarathy, Talfan Evans, Lucas Beyer, Ye~Xia, Basil Mustafa, Olivier H{\'e}naff, Jeremiah Harmsen, Andreas Steiner, and Xiaohua Zhai. 2025.
\newblock ``{SigLIP 2}: Multilingual Vision-Language Encoders with Improved Semantic Understanding, Localization, and Dense Features.''   {\it arXiv preprint arXiv:2502.14786\/}.
\newblock \url{https://arxiv.org/abs/2502.14786}.

\bibitem[\protect\citeauthoryear{{TypeSafe AI}}{{TypeSafe AI}}{2026}]{typesafe2026jev}
{TypeSafe AI} 2026.
\newblock ``Introducing System One Models \& {Jev}.''   Official announcement, September 15.
\newblock \url{https://typesafe.ai/blog/introducing-system-one-models-and-jev}.

\bibitem[\protect\citeauthoryear{{StartLux Labs}}{{StartLux Labs}}{2026}]{startlux2026}
{StartLux Labs} 2026.
\newblock ``{StartLux-Decision-v1}: Open Typed Decision Models.''   Official software and model documentation.
\newblock \url{https://github.com/StartLuxLabs/StartLux-Decision/blob/main/v1/README.md}.

\bibitem[\protect\citeauthoryear{Hu, Shen, Wallis, Allen-Zhu, Li, Wang, Wang, and Chen}{Hu et~al.}{2021}]{hu2021lora}
Edward~J. Hu, Yelong Shen, Phillip Wallis, Zeyuan Allen-Zhu, Yuanzhi Li, Shean Wang, Lu~Wang, and Weizhu Chen. 2021.
\newblock ``{LoRA}: Low-Rank Adaptation of Large Language Models.''   {\it arXiv preprint arXiv:2106.09685\/}.
\newblock \url{https://arxiv.org/abs/2106.09685}.

\bibitem[\protect\citeauthoryear{He, Zhang, Ren, and Sun}{He et~al.}{2016}]{he2016resnet}
Kaiming He, Xiangyu Zhang, Shaoqing Ren, and Jian Sun. 2016.
\newblock ``Deep Residual Learning for Image Recognition.''   In {\it Proceedings of the IEEE Conference on Computer Vision and Pattern Recognition}.
\newblock \url{https://arxiv.org/abs/1512.03385}.

\bibitem[\protect\citeauthoryear{Tan and Le}{Tan and Le}{2019}]{tan2019efficientnet}
Mingxing Tan and Quoc Le. 2019.
\newblock ``{EfficientNet}: Rethinking Model Scaling for Convolutional Neural Networks.''   In {\it Proceedings of the 36th International Conference on Machine Learning}, Vol~97 of {\it Proceedings of Machine Learning Research},   6105--6114.
\newblock \url{https://proceedings.mlr.press/v97/tan19a.html}.

\bibitem[\protect\citeauthoryear{Liu, Lin, Cao, Hu, Wei, Zhang, Lin, and Guo}{Liu et~al.}{2021}]{liu2021swin}
Ze~Liu, Yutong Lin, Yue Cao, Han Hu, Yixuan Wei, Zheng Zhang, Stephen Lin, and Baining Guo. 2021.
\newblock ``{Swin Transformer}: Hierarchical Vision Transformer Using Shifted Windows.''   In {\it Proceedings of the IEEE/CVF International Conference on Computer Vision}.
\newblock \url{https://arxiv.org/abs/2103.14030}.

\bibitem[\protect\citeauthoryear{Liu, Mao, Wu, Feichtenhofer, Darrell, and Xie}{Liu et~al.}{2022}]{liu2022convnext}
Zhuang Liu, Hanzi Mao, Chao-Yuan Wu, Christoph Feichtenhofer, Trevor Darrell, and Saining Xie. 2022.
\newblock ``A {ConvNet} for the 2020s.''   In {\it Proceedings of the IEEE/CVF Conference on Computer Vision and Pattern Recognition}.
\newblock \url{https://arxiv.org/abs/2201.03545}.

\bibitem[\protect\citeauthoryear{Loshchilov and Hutter}{Loshchilov and Hutter}{2019}]{loshchilov2019decoupled}
Ilya Loshchilov and Frank Hutter. 2019.
\newblock ``Decoupled Weight Decay Regularization.''   In {\it International Conference on Learning Representations}.
\newblock \url{https://arxiv.org/abs/1711.05101}.

\end{thebibliography}
\end{document}